\documentclass[letterpaper]{article} 
\usepackage{aaai2026}  
\usepackage{times}  
\usepackage{helvet}  
\usepackage{courier}  
\usepackage[hyphens]{url}  
\usepackage{graphicx} 
\usepackage{natbib}  
\usepackage{caption} 
\usepackage{algorithm}
\usepackage{algorithmic}
\usepackage{amsmath} 
\usepackage{amssymb}
\usepackage{booktabs}
\usepackage{multirow}
\usepackage{xcolor}
\usepackage{svg}

\usepackage{newfloat}
\usepackage{listings}
\DeclareCaptionStyle{ruled}{labelfont=normalfont,labelsep=colon,strut=off} 
\floatstyle{ruled}
\newfloat{listing}{tb}{lst}{}
\floatname{listing}{Listing}
\title{SD-PSFNet: Sequential and Dynamic Point Spread Function Network for Image Deraining}
\author{
    Jiayu Wang\textsuperscript{\rm 1}\equalcontrib, 
    Haoyu Bian\textsuperscript{\rm 2,\rm 3}\equalcontrib,
    Haoran Sun\textsuperscript{\rm 1}, 
    Shaoning Zeng\textsuperscript{\rm 2,\rm 3}\thanks{Corresponding Author}
}
\affiliations{
    \textsuperscript{\rm 1}School of Information and Software Engineering, University of Electronic Science and Technology of China, Chengdu, China\\
    \textsuperscript{\rm 2}Yangtze Delta Region Institute (Huzhou), University of Electronic Science and Technology of China, Huzhou, China\\
    \textsuperscript{\rm 3}School of Computer Science and Engineering, University of Electronic Science and Technology of China, Chengdu, China\\
   aster20060626@gmail.com, haoyubian04@gmail.com, 2022090916002@std.uestc.edu.cn, zsn@outlook.com
}

\usepackage{bibentry}

\begin{document}

\maketitle

\begin{abstract}
Image deraining is crucial for vision applications but is challenged by the complex multi-scale physics of rain and its coupling with scenes. To address this challenge, a novel approach inspired by multi-stage image restoration is proposed, incorporating Point Spread Function (PSF) mechanisms to reveal the image degradation process while combining dynamic physical modeling with sequential feature fusion transfer, named SD-PSFNet. Specifically, SD-PSFNet employs a sequential restoration architecture with three cascaded stages, allowing multiple dynamic evaluations and refinements of the degradation process estimation. The network utilizes components with learned PSF mechanisms to dynamically simulate rain streak optics, enabling effective rain-background separation while progressively enhancing outputs through novel PSF components at each stage. Additionally, SD-PSFNet incorporates adaptive gated fusion for optimal cross-stage feature integration, enabling sequential refinement from coarse rain removal to fine detail restoration. Our model achieves state-of-the-art PSNR/SSIM metrics on Rain100H (33.12dB/0.9371), RealRain-1k-L (42.28dB/0.9872), and RealRain-1k-H (41.08dB/0.9838). In summary, SD-PSFNet demonstrates excellent capability in complex scenes and dense rainfall conditions, providing a new physics-aware approach to image deraining.
\end{abstract}

\begin{links}
    \link{Code}{https://github.com/Aster-1024/SD-PSFNet}
\end{links}

\section{Introduction}
Image deraining is a fundamental restoration task aiming to recover rain-free images from rainy ones. This task is crucial for downstream applications \cite{su2023survey} such as object detection \cite{wang2022end}, autonomous driving \cite{sun2019convolutional}, and surveillance systems \cite{dey2020single}, as rain-degraded images suffer from reduced visibility, decreased contrast, and detail loss that significantly impact visual processing performance. In recent years, deep learning-based deraining methods have made significant progress \cite{wang2020single}. From early CNN-based models to approaches incorporating Transformer architectures such as Restormer \cite{zamir2022restormer} and Uformer \cite{wang2022uformer}, as well as GAN-based techniques like SSCGAN \cite{shao2021selective} and LFI-based \cite{ding2021rain}, these approaches have demonstrated remarkable capabilities in image deraining tasks.

The primary challenges in image deraining arise from the complexity, multi-scale nature, and high coupling of raindrop distributions with background scenes. Most existing methods primarily focus on learning mapping relationships from data, overlooking the physical and optical properties of raindrop formation \cite{yang2020single}. This neglect leads to an inability to dynamically adjust to different rain patterns, resulting in insufficient deraining and lack of explainability. Although Transformer-based architectures, GAN-based models, and diffusion models demonstrate superior performance in image restoration tasks, their massive parameter counts and computational overhead hinder deployment in real-time applications \cite{chen2025towards}. Additionally, existing methods attempt to incorporate domain knowledge through physics-aware approaches, such as leveraging rain streaks' frequency domain characteristics as inductive bias. However, traditional explicit physical modeling relies on fixed prior assumptions, such as static PSF templates that struggle to capture the diversity of rain degradation, leading to overlooking the multi-scale distribution of rain streaks \cite{zhu2020physical} and optical differences of raindrops.

To address these issues, we refocus our attention on CNN architecture and propose a sequential modeling-based deraining network featuring a novel physics-aware mechanism that relies on multi-scale, data-driven dynamic Point Spread Function (PSF) \cite{rossmann1969point} prediction. Based on this mechanism, we propose SD-PSFNet, constructing multiple innovative PSF-related modules integrated into a powerful multi-stage architecture, inspired by successful progressive restoration paradigms like MPRNet \cite{zamir2021multi}. Additionally, we implement an adaptive gating fusion mechanism to achieve sequential modeling, facilitating the transmission and optimization of cross-stage information within the multi-stage architecture.

Experiments show SD-PSFNet achieves state-of-the-art deraining performance across public datasets while maintaining reasonable computational efficiency. Compared to the baseline MPRNet architecture, our novel physics-aware design delivers a significant 5.04dB PSNR improvement (13.5\% gain) on RealRain-1k-L, with particular effectiveness in complex scenes and dense rain conditions. The key contributions are:

\begin{itemize}
    \item \textbf{A Novel PSF-Integrated Deraining Approach}: Multiple integrated PSF mechanism modules dynamically simulate raindrops' optical effects, enhancing representation of complex rain patterns.

    \item \textbf{Sequential Model Design}: Multi-level gated fusion strategies enable full transmission of physical prior information across multiple stages, repeatedly and dynamically refining the estimation of image degradation.
    
    \item \textbf{Extensive Experimental Evaluation}: Multiple benchmark evaluations, including comparative tests and ablation studies, confirm our method's superior performance.
\end{itemize}

\section{Related Work}
\subsection{Image Deraining}
Image deraining aims to recover rain-free images from rainy ones, representing a specific instance of reversing image degradation. Traditional methods enhanced interpretability by modeling rain's optical properties through image decomposition \cite{kang2011automatic} and frequency domain analysis \cite{zheng2013single}. However, these approaches struggle with complex scenes, causing over-smoothing or residual rain streaks.

Deep learning has emerged as the primary approach for image deraining, evolving from CNN-based local feature extraction to global modeling and generative techniques. Early CNNs like DANet \cite{jiang2022danet} and DerainRLNet \cite{chen2021robust} processed rain hierarchically but suffered from limited receptive fields. Later, multi-scale networks such as RESCAN \cite{li2018recurrent} and DID-MDN \cite{zhang2018density} along with physically-guided networks like KGCNN \cite{wang2020rain} enhanced adaptability through multi-resolution feature fusion and motion blur modeling. Restormer \cite{zamir2022restormer} applied Transformer \cite{vaswani2017attention} to low-level vision, achieving linear-complexity long-range dependency modeling via cross-channel self-attention. Generative approaches including Attentive GAN \cite{qian2018attentive} produced realistic rain-free images through adversarial training and attention mechanisms, though facing mode collapse issues. Semi-supervised and unsupervised methods such as UD-GAN \cite{jin2019unsupervised} and DerainCycleGAN \cite{wei2021deraincyclegan} addressed the scarcity of real rain annotations by leveraging unpaired data, while DCDGAN \cite{chen2022unpaired} utilized diffusion models for enhanced realism. However, these methods still lack multi-scale physics-aware modeling and struggle to balance computational efficiency with deraining performance.

\subsection{Physics-Aware Image Restoration}
Physics-aware approaches are widely used in image restoration, integrating optical imaging's physical mechanisms into models \cite{xu2023physics}. Fusion strategies combining explicit physical modeling with data-driven deep learning approaches have become mainstream methods. For example, ReDNet \cite{wu2025restoration} constructed an optically-guided Transformer network that precisely separates stains from background information. In applications focusing on rainy scenes, the ReDT-Det \cite{chen2025retinex} network employs Illumination Fusion Differential Transformer Blocks (IFDTB) to suppress noise and restore details in dark regions, significantly enhancing vehicle detection capabilities in nighttime rainy and foggy conditions. Similarly, researchers have proposed a Dual-Scale Prior (DSP) \cite{yuan2025image} model that combines physical priors such as non-local self-similarity with the powerful representational capabilities of pre-trained deep denoising networks, achieving robust removal of degradation factors like rain streaks. However, multi-scale rain streak removal still requires further exploration and enhancement.

\begin{figure*}[h]
  \centering
  \includegraphics[width=\linewidth]{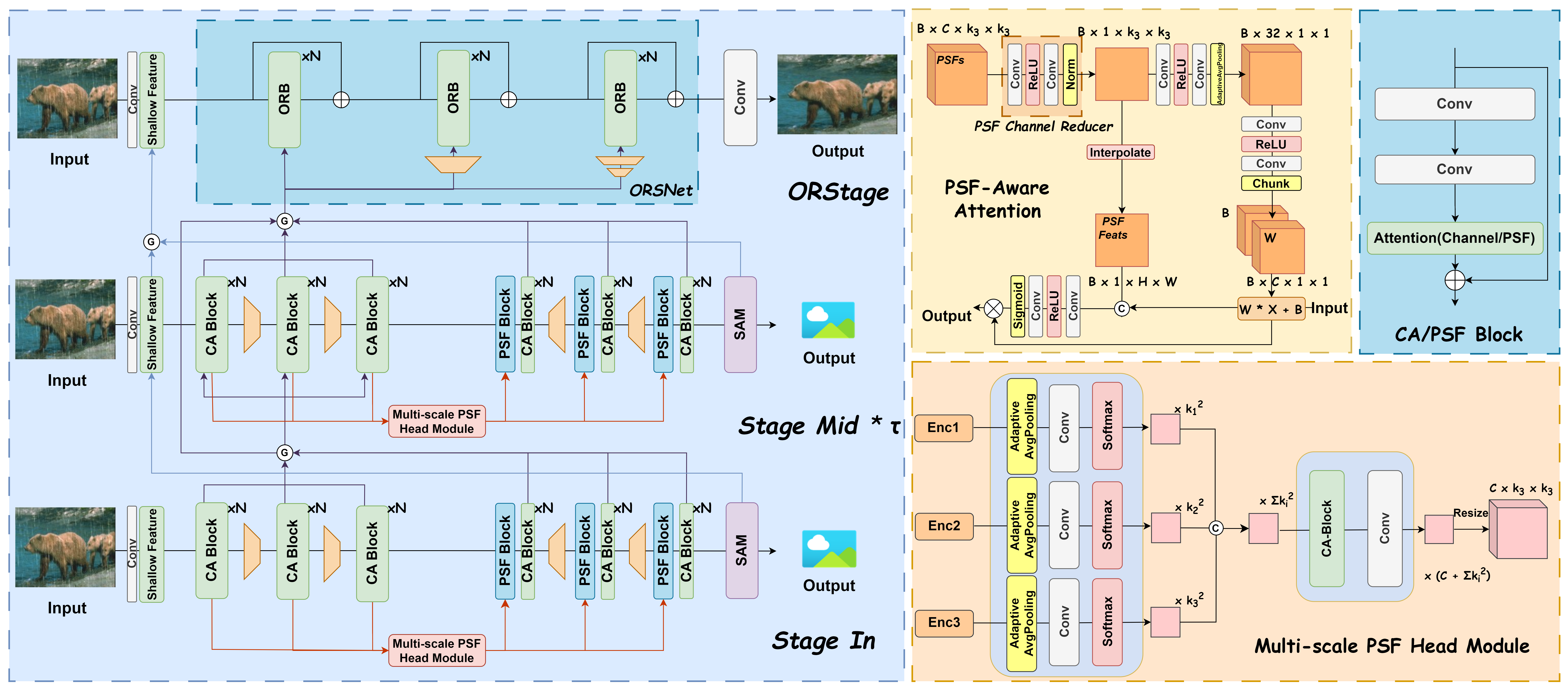}
  \caption{Overview of the SD-PSFNet. The left panel shows the serialized overall restoration framework of SD-PSFNet, including three stages: input Stage In, multiple Stage Mid, and final restoration output ORStage. The right panels detail our key components: PSF Channel Reducer, PSF-Aware Attention, CA/PSF Block, and Multi-Scale PSF Head Module. The network integrates multi-scale dynamic PSF mechanisms for modeling, effectively handling multi-scale rain removal. Input images undergo serialized restoration through three stages with specialized PSF-aware processing.}
  \label{fig:first}
\end{figure*}

\section{Method}
This section details SD-PSFNet's implementation, including theoretical validation of PSF and the innovative module and framework design.

\subsection{Dynamic PSF Mechanism}
Image degradation processes can typically be modeled as linear space-invariant systems using Point Spread Function (PSF), but real rainy image degradation is more complex, with traditional models struggling to accurately characterize spatially varying degradation fields \( K(x, y) \). We transform PSF representation from static presets to data-driven learnable forms. Specifically, we propose that any spatially variable degradation function \( K(x, y) \) can be approximated through content-adaptive linear combinations of a learnable degradation pattern dictionary \( \{k_j\}_{j=1}^{K_c} \):
\[
K(x,y) \approx \sum_{j=1}^{K_c} w_j(x,y) \cdot k_j
\]
where $k_j$ represents learnable 2D convolution kernels capturing basic degradation patterns, collectively forming a dictionary representing various rain streak degradation types. Since rain degradation is dominated by finite dimensions like direction and density, a comprehensive pattern dictionary can effectively cover its variation range. In our implementation, neural networks achieve adaptive mapping from image features to combination weights with added physical constraints.

\subsection{Dynamic Physics-Aware Modules}
Point Spread Function (PSF) characterizes imaging systems' response to point light sources, critical for understanding image degradation. Image deraining essentially reverses complex, spatially varying degradation processes. Unlike traditional approaches using static PSFs, SD-PSFNet dynamically adapts to each image's unique degradation patterns, effectively overcoming conventional model limitations.

\textbf{Multi-Scale PSF Head} synthesizes degradation information from multi-level encoder features, dynamically performing adaptive PSF modeling for various degradation patterns across different image regions, overcoming the limitations of traditional methods with single-scale or predefined degradation models.

For a single-scale PSF prediction head, we process encoder features \(F_i \in \mathbb{R}^{B \times C_i \times H_i \times W_i}\) to generate the PSF representation:
\begin{equation}
\begin{split}
P_i^{raw} = \mathcal{W}_i(\mathcal{AP}(F_i)) \in \mathbb{R}^{B \times k_i^2 \times 1 \times 1}
\end{split}
\end{equation}
Here, \(\mathcal{AP}\) is adaptive average pooling, \(B\) is batch size, and \(\mathcal{W}_i\) is a 1×1 convolutional layer mapping \(C_i\)-dimensional features to \(k_i^2\) dimensions, functionally equivalent to a multi-layer perceptron (MLP). Subsequently, Softmax is applied along the second dimension (kernel dimension) to normalize \(P_i^{\text{raw}}\) into a valid probability distribution, yielding the flattened weight vector \(P_i^{\text{flat}} \in \mathbb{R}^{B \times k_i^2 \times 1 \times 1}\).

In the multi-scale scenario, we employ multiple prediction heads with specific sizes (3×3, 5×5, and 7×7) to extract complementary degradation information from encoder features $\{F_1, F_2, ..., F_n\}$. These designed heads focus on high-frequency detail degradation, balanced local and global information, and macroscopic low-frequency patterns respectively. The flattened PSFs are concatenated along the channel dimension to form $P_{concat}^{flat}$. A channel attention block (CAB) \cite{zhang2018image} then adaptively fuses these multi-scale PSF representations into $P_{att}^{flat}$. This fused representation is processed by a 1×1 convolutional layer that projects it into $P_{refined}^{flat} \in \mathbb{R}^{B \times (K_c \cdot K \cdot K) \times 1 \times 1}$, where $K_c$ is the number of channels and $K$ is the spatial size of the output PSF. Finally, $P_{refined}^{flat}$ is reshaped into the multi-channel PSF representation $P_{multi} \in \mathbb{R}^{B \times K_c \times K \times K}$.

To ensure each channel of $P_{multi}$ forms a valid probability distribution kernel, preventing unrealistic amplification or attenuation of background pixels and simulating energy conservation, we enforce spatial normalization:
\begin{equation}
\begin{split}
\sum_{x=1}^{K} \sum_{y=1}^{K} P_{multi}(b,c,x,y) = 1,\\\ \quad \forall b \in \{1,...,B\}, \forall c \in \{1,...,K_c\}
\end{split}
\end{equation}

Each channel of $P_{multi}$ represents a principal component in the degradation space. Through channel attention fusion, the model learns to integrate complementary multi-scale information from individual heads, collectively forming a complete representation of the degradation patterns.

\textbf{PSF Block (PSFB)} guides image deraining using PSF features from the \textbf{Multi-Scale PSF Head}. It adaptively adjusts input features through dynamic modulation and incorporates physical degradation models, outperforming traditional attention methods in restoring spatially varying defects. Its refinement process is driven by the \textbf{PSF Channel Reducer} and \textbf{PSF-Aware Attention} components.

\textbf{PSF Channel Reducer} enables information compression by transforming multi-channel PSF from the Multi-Scale PSF Head into a single-channel representation. This module preserves key degradation information through adaptive channel attention. When output channels are fewer than the input, it applies direct mapping using 1×1 convolution. Conversely, it first compresses to \(\frac{1}{4}\) of the original channels before mapping to the target count. The compressed PSF then undergoes spatial normalization to ensure its elements sum to 1. This approach maintains the physical properties of the PSF while reducing computational complexity.

\textbf{PSF-Aware Attention} adaptively adjusts feature representations using estimated PSF degradation information, achieving physics-guided feature enhancement. This module integrates PSF encoding with input features through a dual-path mechanism of channel modulation and spatial attention. As shown in Figure~\ref{fig:first}, for multi-channel PSF ($K_c > 1$), channels are first compressed via the PSF Channel Reducer. A dedicated PSF Generator extracts features at multiple stages to predict degradation kernels at different scales. These predictions are normalized via softmax and fused through channel attention mechanisms to produce a multi-channel PSF representation. The PSF Encoder, integrated within the attention module, processes this information through convolutional layers and adaptive pooling to extract a compact feature representation \(P_{feat}\). This generates channel modulation parameters \(\gamma\) and \(\beta\):

\begin{equation}
\begin{split}
[\gamma, \beta] = \mathcal{F}_{channel}(P_{feat}) \in \mathbb{R}^{B \times C \times 2}
\end{split}
\end{equation}
where \(\mathcal{F}_{channel}\) is a fully-connected network that maps PSF features to modulation parameters matching the input feature channels. Channel modulation enhances features:

\begin{equation}
\begin{split}
x_{mod} = x \odot \gamma + \beta \in \mathbb{R}^{B \times C \times H \times W}
\end{split}
\end{equation}
where \(\odot\) denotes element-wise multiplication, with modulation parameters broadcast across all spatial positions. The spatial attention branch upsamples the single-channel PSF to match the input feature's dimensions and combines it with the modulated features to generate spatial attention weights, highlighting regions affected by different degradation levels. Finally, channel-modulated features are multiplied by these spatial attention weights to produce output features.

\subsection{Sequential Design and Enhanced Feature Fusion}
In traditional cascade networks, shallow detail loss and deep semantic gaps often occur. This paper employs a cross-stage gating mechanism to achieve feature transfer through adaptive weights, expressed as:

\begin{equation}
\begin{split}
\mathbf{F}^{(t)} = G_{\theta}\left(\mathbf{F}_{\text{current}}, \mathbf{F}_{\text{prev}}\right) \odot \mathbf{F}_{\text{current}} +\\ \left(1 - G_{\theta}\left(\mathbf{F}_{\text{current}}, \mathbf{F}_{\text{prev}}\right)\right) \odot \mathbf{F}_{\text{prev}}
\end{split}
\end{equation}

Here, the gating weight \(G_{\theta} \in [0,1]\) is learned from the current feature \(\mathbf{F}_{\text{current}}\) and the historical feature \(\mathbf{F}_{\text{prev}}\). The gating mechanism dynamically fuses features from different sources through an adaptive weighting process. It concatenates the input features and applies global average pooling to capture their overall characteristics. These pooled features pass through convolutional layers to generate fusion weights, determining the contribution of each input feature to the final output, effectively emphasizing more informative features while suppressing less useful ones.

This mechanism operates at three key points: at stage inputs where adaptive gating fuses the current input with previous stage features after shallow feature extraction, at encoder levels through integration of current outputs with historical features, and via enhanced Cross-Stage Feature Fusion (CSFF). This improved CSFF efficiently routes information between network stages through dual adaptive gating units. Here, $\tau$ represents the number of middle stages, $I_{\text{in}}$ is the degraded input, and $\hat{I}$ is the restored output. Each stage outputs an intermediate restoration result $I_i$, cross-scale features $O_i$ via CSFF, and feature mappings $H_i$. The feature extraction module incorporates CSFF through:

\begin{equation}
\begin{split}
\mathcal{F}_{\text{shallow}}(I_{\text{in}}, H) = \mathcal{F}_{\text{gate}}(\mathcal{F}_{\text{conv+CAB}}(I_{\text{in}}), H)
\end{split}
\end{equation}

In the CSFF process, encoder and decoder features from the current stage first pass through an initial gating unit to determine their relative importance. After convolution, these processed features are fed into a second gating unit where they're adaptively fused with features from the previous stage. This fusion undergoes another convolution to generate output features for guiding the next stage. This dual-gating mechanism provides precise control over cross-stage information flow, preserving both low-level details and high-level semantic information throughout the network.

This approach ensures that texture details and semantic information from earlier stages contribute to the refinement in subsequent stages, effectively addressing attenuation and gaps in feature transfer within multi-stage networks.

In the implementation within the original-resolution subnetwork (ORSNet) \cite{zamir2021multi}, unlike the original method, the SD-PSFNet design incorporates multi-scale information at specific processing depths, rather than just at the input and output boundaries. Each original resolution block (ORB) receives enriched features from both the main path and transformed side information, enabling fine-grained texture reconstruction at the original resolution.

This approach ensures that texture details and semantic information from earlier stages contribute to the refinement in subsequent stages through gating weights, effectively addressing attenuation and gaps in feature transfer within multi-stage networks.

\begin{table*}[h]
\centering
\small
\begin{tabular}{l|c|cc|cc|cc|cc}
\toprule
\multicolumn{2}{c|}{Datasets} & \multicolumn{2}{c|}{Rain100L} & \multicolumn{2}{c|}{Rain100H} & \multicolumn{2}{c|}{Realrain-1k-L} & \multicolumn{2}{c}{Realrain-1k-H} \\
Method & Params(M) & PSNR(dB) & SSIM & PSNR(dB) & SSIM & PSNR(dB) & SSIM & PSNR(dB) & SSIM \\
\midrule
\multicolumn{10}{c}{\textbf{CNN-based Methods}} \\
\midrule
DerainNet & 0.75 & 27.03 & 0.8841 & 14.92 & 0.5923 & 27.09 & 0.925 & 22.88 & 0.8886 \\
DDN & 0.06 & 32.38 & 0.9259 & 24.64 & 0.849 & 31.18 & 0.9172 & 29.17 & 0.8783 \\
PreNet & 0.28 & 37.48 & 0.979 & 29.46 & 0.899 & 33.01 & 0.944 & 30.88 & 0.9108 \\
SPANet & 0.28 & 35.33 & 0.969 & 25.11 & 0.827 & 30.43 & 0.947 & 25.76 & 0.9095 \\
MPRNet & 3.64 & 36.4 & 0.965 & 30.41 & 0.89 & 36.29 & 0.972 & 34.74 & 0.964 \\
NAFNet & 17.11 & 37 & 0.978 & 29.66 & 0.9 & 38.8 & 0.986 & 36.11 & 0.976 \\
HINet & 88.67 & 37.28 & 0.97 & 30.65 & 0.894 & 41.98$^\dag$ & 0.9869$^\dag$ & 40.82$^\dag$ & 0.983$^\dag$ \\
M3SNet & 16.70 & 40.04 & 0.985 & 30.64 & 0.982 & 41.55 & 0.9852 & 40.01 & 0.979 \\
\textbf{Ours} & \textbf{9.63} & \textbf{41.47}$^\dag$ & \textbf{0.9896}$^\dag$ & \textbf{33.12}$^*$ & \textbf{0.9371}$^*$ & \textbf{42.28}$^*$ & \textbf{0.9872}$^*$ & \textbf{41.08}$^*$ & \textbf{0.9838}$^*$ \\
\midrule
\multicolumn{10}{c}{\textbf{Transformer-based Methods}} \\
\midrule
Restormer & 26.10 & 38.99 & 0.978 & 31.46 & 0.904 & 40.9 & 0.9849 & 39.57 & 0.9812 \\
PromptIR & 32.11 & 38.34 & 0.983 & 28.69 & 0.877 & 36.99 & 0.973 & 33.61 & 0.953 \\
DRSFormer & 33.66 & 41.32 & 0.9887 & 32.07 & 0.9316 & 41.52 & 0.9812 & 40.21 & 0.9824 \\
NeRD-Rain-S & 10.53 & 42$^*$ & 0.99$^*$ & 32.86$^\dag$ & 0.932$^\dag$ & 38.64 & 0.979 & 36.69 & 0.97 \\
\bottomrule
\end{tabular}
\caption{Performance comparison (PSNR/SSIM) of deraining methods on synthetic and real-world datasets. $*$ and $^\dag$ represent the methods ranked first and second in performance metrics on the corresponding dataset. The suffixes L and H represent "Light" and "Heavy" rainfall.}
\label{tab:comparison}
\end{table*}

\subsection{Overview of SD-PSFNet}
SD-PSFNet employs a multi-stage sequential restoration strategy inspired by sequential modeling approaches like LSTM \cite{hochreiter1997long} and multi-stage image restoration models such as MPRNet. Stage-wise image restoration, as a common approach, allows models to repeatedly re-evaluate and refine their estimates of the degradation process while being conducive to module embedding. Notably, although we use designs similar to MPRNet and related module references, such as Supervised Attention Module (SAM) and ORSNet, our approach differs from MPRNet's spatial division strategy (e.g., partitioning images into quarters, half-size, etc.). Instead, we prefer to view SD-PSFNet as a serialized restoration strategy where each stage processes the complete image and dynamically predicts and utilizes PSF features at different resolutions through the multi-scale downsampling structure within UNet.

Our architecture consists of Stage In, Stage Mid, and Original Resolution Stage (ORStage) with enhanced physical information transmission between stages. The overall network is expressed as:
\begin{equation}
\begin{split}
\hat{I} = \mathcal{F}_{\text{ORS}}(I_{\text{in}}, H_{\text{mid}}, O_{\text{mid}})
\end{split}
\end{equation}
Intermediate feature representation $H_{\text{i}}$ and cross-stage features $O_{\text{i}}$ are generated through:
\begin{equation}
\begin{split}
\{I_1, H_1, O_1\} = \mathcal{F}_{\text{StageIn}}(I_{\text{in}})
\end{split}
\end{equation}
\begin{equation}
\begin{split}
\{I_i, H_i, O_i\} = \mathcal{F}_{\text{StageMid}_i}(I_{\text{in}}, H_{i-1}, O_{i-1})
\end{split}
\end{equation}
where $i$ ranges from 2 to $\tau$+1, $\tau$ represents the number of middle stages, $I_{\text{in}}$ is the degraded input, and $\hat{I}$ is the restored output. Each stage employs an encoder-decoder structure with PSF estimation based on encoder-extracted multi-scale features, upgrading traditional Channel Attention Blocks (CAB) to PSF Blocks (PSFB):
\begin{equation}
\begin{split}
X_{\text{out}}^{\text{PSFB}} = X_{\text{in}} + \mathcal{F}_{\text{PSFA}}(\mathcal{F}_{\text{body}}(X_{\text{in}}), \text{PSF})
\end{split}
\end{equation}
Where $\mathcal{F}_{\text{body}}$ consists of two convolutional layers with activation functions, and $\mathcal{F}_{\text{PSFA}}$ represents the PSF-Aware Attention that adapts the decoding process based on degradation characteristics. SD-PSFNet includes components from MPRNet \cite{zamir2021multi}, including the SAM, which won't be elaborated on here.

\subsection{Loss Function}
The model employs a hybrid loss function that jointly optimizes PSF estimation and physics-guided restoration:

\begin{equation}
\begin{split}
\mathcal{L}_{total} = \sum_{s=1}^{\tau+1}( \mathcal{L}_{char}(\mathbf{I}_s, \mathbf{T}) + \alpha_1\mathcal{L}_{edge}(\mathbf{I}_s, \mathbf{T}) \\+ \alpha_2\mathcal{L}_{freq}(\mathbf{I}_s, \mathbf{T}))
\end{split}
\end{equation}

The Charbonnier loss ($\mathcal{L}_{char}$) provides pixel-wise supervision, edge-aware loss ($\mathcal{L}_{edge}$) preserves high-frequency details, and frequency domain loss ($\mathcal{L}_{freq}$) aligns with PSF characteristics. With $\alpha_1=0.05$, $\alpha_2=0.01$, and $\mathbf{I}_s$ representing stage outputs against ground truth $\mathbf{T}$, this approach maintains pixel accuracy while respecting physical degradation constraints for effective PSF-guided restoration.


\section{Experiments}
\subsection{Datasets and Benchmarks} 
Four benchmark datasets are introduced for the main comparative experiments, including synthetic datasets: Rain100L \cite{yang2019joint} and Rain100H \cite{yang2019joint}, and real-world datasets: RealRain-1K-L \cite{li2022toward}, RealRain-1K-H \cite{li2022toward}. What's more, SPA-data \cite{wang2019spatial}, Rain13K \cite{zamir2021multi}, and Rain200L \cite{yang2017deep} which are only incorporated in the Generalization Results section. To evaluate the proposed method, several state-of-the-art methods are selected. We compare our method with CNN-based methods (DerainNet \cite{fu2017clearing}, DDN \cite{fu2017removing}, PreNet \cite{ren2019progressive}, SPANet \cite{wang2019spatial}, MPRNet \cite{zamir2021multi}, NAFNet \cite{chen2022simple}, HINet \cite{chen2021hinet}, M3SNet\cite{gao2023mountain}), Transformer-based methods (Restormer \cite{zamir2022restormer}, PromptIR \cite{potlapalli2023promptir}, DRSFormer \cite{chen2023learning}, NeRD-Rain-S \cite{chen2024bidirectional}). The method employs supervised training, thus excluding comparisons with unsupervised GAN-based deraining approaches.

\begin{figure*}[h]
  \centering
  \includegraphics[width=\linewidth]{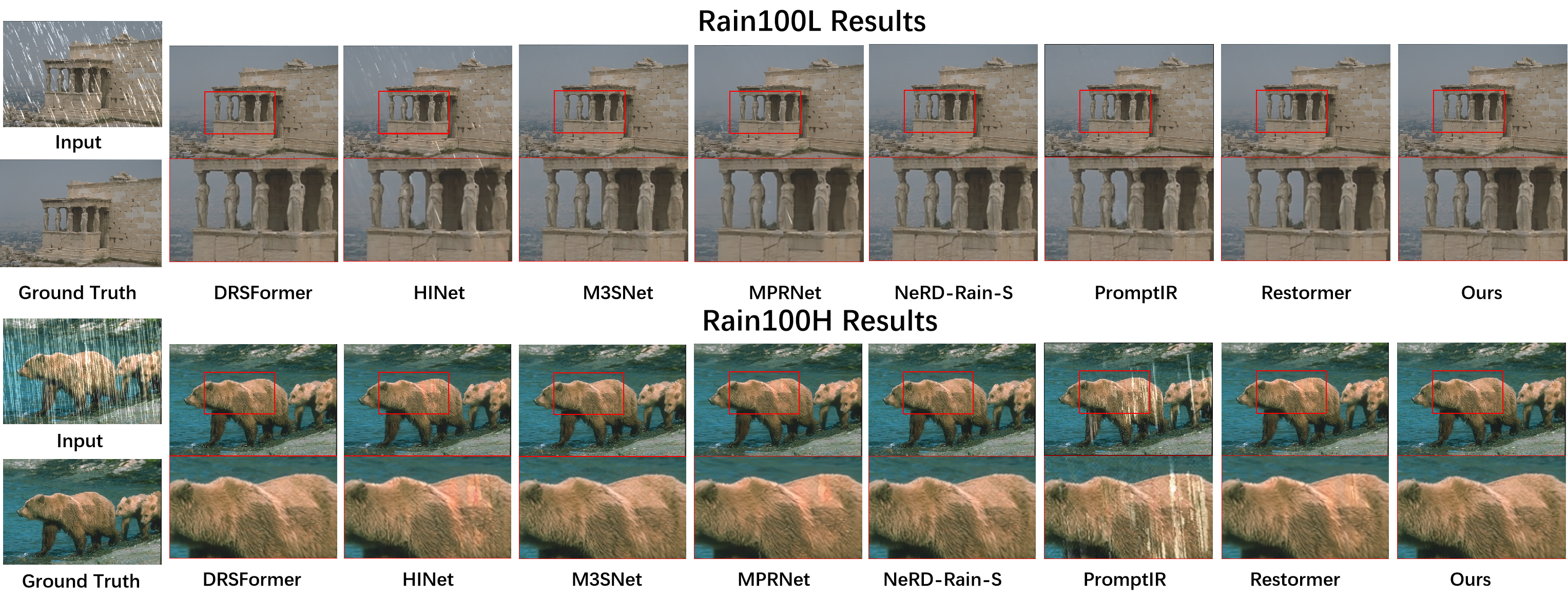}
  \caption{Qualitative deraining performance comparisons on Rain100L \cite{yang2019joint}and Rain100H \cite{yang2019joint}datasets. Our SD-PSFNet achieves competitive visual results comparable to the other SOTA method.}
  \label{fig:result}
\end{figure*}

\begin{table*}[h]
\centering
\begin{tabular}{c|cc|cc|cc|cc|cc}
\hline
\multirow{2}{*}{Train $\downarrow$ Test $\rightarrow$} & \multicolumn{2}{c|}{Realrain-1k-L} & \multicolumn{2}{c|}{Realrain-1k-H} & \multicolumn{2}{c|}{Rain100L} & \multicolumn{2}{c|}{Rain100H} & \multicolumn{2}{c}{SPA-Data} \\
 & PSNR & SSIM & PSNR & SSIM & PSNR & SSIM & PSNR & SSIM & PSNR & SSIM \\
\hline
RealRain-1k-L & \textbf{42.28} & \textbf{0.9872} & 40.70 & 0.9829 & 26.93 & 0.8377 & 14.34 & 0.3907 & \textbf{34.89} & \textbf{0.9617} \\
RealRain-1k-H & 41.14 & 0.9855 & \textbf{41.08} & \textbf{0.9838} & 23.61 & 0.7949 & 16.79 & 0.4719 & 33.23 & 0.9552 \\
Rain100L & 27.54 & 0.8831 & 23.82 & 0.7963 & \textbf{41.47} & \textbf{0.9896} & 18.79 & 0.5721 & 32.97 & 0.9206 \\
Rain100H & 26.98 & 0.8733 & 24.12 & 0.8106 & 38.07 & 0.9797 & \textbf{33.12} & \textbf{0.9371} & 31.93 & 0.9121 \\
\hline
\end{tabular}
\caption{Cross-dataset evaluation results on synthetic and real rain datasets. Each row represents a model trained on a specific dataset, while columns show testing performance on different datasets.}
\label{tab:cross_evaluation_synthetic}
\end{table*}

\subsection{Implementation Details and Metrics}
Our model is implemented using the PyTorch framework with $\tau$=3 stages and PSF channel $K_c$=40 for main results and trained for 2000 epochs in Table~\ref{tab:comparison}. We employ AdamW optimizer with an initial learning rate of 1e-4, incorporating a 3-epoch linear warmup followed by cosine annealing to 1e-6. Mixed-precision training (FP16) accelerates computation while gradient clipping (max norm 2.0) stabilizes convergence. During training, rain-free images and ground truth are aligned and cropped to 128×128 patches, normalized using ImageNet mean-variance statistics, and augmented with random flipping and geometric transformations. For testing, we only apply minimal reflection padding to maintain whole-image inference. For quality evaluation of derained images, we employ PSNR and SSIM metrics. During training and validation, we calculate PSNR only on the Y channel (luminance component in YCrCb), while during testing, we compute both PSNR and SSIM in RGB space. All experiments are conducted from scratch on a machine with one NVIDIA GeForce RTX 4090 GPU.

\subsection{Main Deraining Results}
\begin{figure*}[h]
  \centering
  \includegraphics[width=\linewidth]{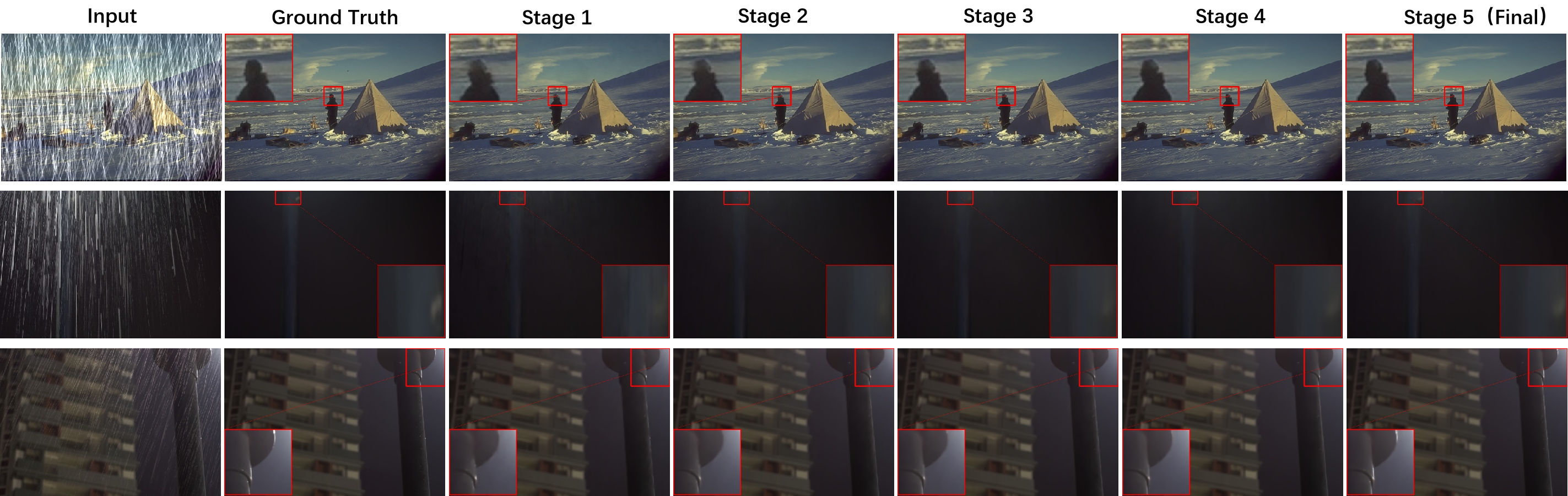}
  \caption{Comparison of restoration results across different stages on three datasets (from top to bottom): Rain100H \cite{yang2019joint} and two RealRain-1k-H \cite{li2022toward} samples. Stage 1 represents Stage In, Stages 2-4 correspond to Stage Mid, and Stage 5 shows the final restoration from ORStage.}
  \label{fig:result2}
\end{figure*}

As shown in Table~\ref{tab:comparison}, our proposed CNN-based method achieves impressive performance across most evaluation metrics and datasets. For synthetic benchmarks, we obtain 41.47 dB PSNR/0.9896 SSIM on Rain100L and 33.12 dB PSNR/0.9371 SSIM on Rain100H, demonstrating that our CNN architecture can match or even surpass many Transformer-based methods like Restormer and PromptIR. Although NeRD-Rain-S shows slightly better results on Rain100L, our method demonstrates extreme performance on real-world datasets, achieving 42.28 dB PSNR/0.9872 SSIM on RealRain-1k-L and 41.08 dB PSNR/0.9838 SSIM on RealRain-1k-H. Our CNN-based method bridges the performance gap with state-of-the-art Transformer architectures while maintaining CNN efficiency advantages, handling diverse rain patterns in both synthetic and real-world scenarios. Visualizations are shown in Figures~\ref{fig:result} and \ref{fig:result2}.

\subsection{Generalization Results}
To validate the generalization capability of SD-PSFNet and the significant domain gap between artificial and real rainfall scenarios, as shown in Table \ref{tab:cross_evaluation_synthetic}, models trained on the RealRain-1k dataset demonstrate poor transferability to the Rain100 dataset, and vice versa, with severe performance degradation when crossing domain boundaries. Notably, since the Ground Truth in SPA-Data is derived from video-based deraining methods, our method demonstrates good and more stable performance on SPA-Data. This gap represents a fundamental limitation of data-driven approaches, as models learn dataset-specific features rather than universal deraining principles. Table \ref{tab:zero_shot_transfer} demonstrates generalization comparison from synthetic to real datasets, showcasing the generalization capability of various deraining methods when applied to real-world rain scenarios without specific fine-tuning. SD-PSFNet achieves the highest PSNR of 26.98dB and maintains consistent superior performance when trained on the Rain100H, demonstrating that our method effectively captures the essential characteristics of rain streaks and shows high generalization capability compared to other models trained on the larger datasets.

\begin{table}[h]
\centering
\caption{Generalization comparison from synthetic to real datasets on the RealRain-1k-L dataset.}
\label{tab:zero_shot_transfer}
\begin{tabular}{l|c|c}
\hline
\multirow{2}{*}{Method} & \multicolumn{1}{c|}{Metrics} & \multirow{2}{*}{Training Dataset} \\
\cline{2-2}
 & PSNR & \\
\hline
MPRNet & 25.55 & Rain13K \\
Restormer & 26.59 & Rain13K \\
NeRD-Rain-S & 26.67 & Rain200L \\
\textbf{SD-PSFNet} & \textbf{26.98} & \textbf{Rain100H} \\
\hline
\end{tabular}
\end{table}

\begin{table}[h]
\centering
\begin{tabular}{l|cc}
\hline
\multirow{2}{*}{Model Configuration} & \multicolumn{2}{c}{RealRain-1k-L} \\
\cline{2-3}
 & PSNR & SSIM \\
\hline
MPRNet (Baseline) & 37.24dB & 0.9754 \\
- Patch Split & 37.20dB & 0.9755\\
+ Gate & 38.65dB & 0.9773 \\
+ H Inter-stage Updates & 40.41dB & 0.9792 \\
+ Enhanced CSFF & 41.41dB & 0.9850 \\
+ 1-channel PSF & 41.63dB & 0.9855 \\
+ 40-channel PSF (Ours) & 42.28dB & 0.9872 \\
\hline
\end{tabular}
\caption{Ablation study results on RealRain-1k-L dataset. (components added incrementally)}
\label{tab:sum}
\end{table}

\begin{table}[h]
\centering
\small
\begin{tabular}{l c c c c}
\hline
Parameter & PSNR & SSIM & Params (M) & MACs (G) \\
\hline
$\tau=0$ & 40.81dB & 0.9832 & 3.64 & 135.92 \\
$\tau=1$ & 40.86dB & 0.9835 & 5.64 & 170.67 \\
$\tau=2$ & 40.97dB & 0.9843 & 7.63 & 206.97 \\
$\tau=3$ & 41.54dB & 0.9862 & 9.63 & 244.83 \\
\hline
\end{tabular}
\caption{Ablation study on different values of parameter $\tau$.}
\label{tab:parameter}
\end{table}

\subsection{Ablation Studies}
Ablation studies were conducted to evaluate the effectiveness of each proposed component in SD-PSFNet and reveal the size of the model in different numbers of stages.

Incremental analysis on the RealRain-1k-L dataset shows substantial performance gains with each module in Table~\ref{tab:sum}. The baseline, MPRNet with patch-splitting, confirms its advantage over the variant without it. Ablation results indicate that removing patch-splitting leads to decreased performance. The analysis shows significant gains with each module, with patch-splitting removal decreasing performance. The gate mechanism improves performance by 1.45dB, while hierarchical inter-stage updates add 1.76dB, highlighting the value of progressive feature refinement. The enhanced CSFF module contributes an additional 1.0dB, and the 1-channel PSF offers a modest 0.22dB gain. The final architecture achieves optimal performance of 42.28dB PSNR and 0.9872 SSIM, reflecting a substantial 5.04dB improvement over the baseline. These results highlight the meaningful contributions of each proposed component.

Parameter $\tau$ examination over 1000 epochs reveals consistent performance improvements as $\tau$ increases from 0 to 3 in Tabel~\ref{tab:parameter}. Without Stage Mid ($\tau$=0), the model achieves 40.81dB PSNR and 0.9832 SSIM with 3.64M parameters and 135.92G MACs. Increasing to $\tau$=3 delivers optimal performance (41.54dB PSNR, 0.9862 SSIM), demonstrating a significant 0.57dB improvement over the baseline configuration, though at the cost of increased model size (9.63M parameters) and computational complexity (244.83G MACs). Our CNN-based architecture efficiently manages these parameter increases while outperforming similar-sized alternatives. This analysis confirms the importance of multiple Stage Mid phases for enhancing restoration capability while illustrating the associated computational trade-offs.

\section{Conclusion}
This paper proposes a sequential model incorporating dynamic physics-aware Point Spread Function (PSF) mechanisms. The approach integrates innovative multi-scale PSF modules into a multi-stage deraining architecture with multi-level gated fusion mechanisms, enhancing degradation feature modeling while suppressing detail loss through cross-stage feature memory transfer. Experimental results show that SD-PSFNet achieves a better balance between performance and computational efficiency compared to existing deraining methods, providing new insights for physics-aware image restoration.

\section{Acknowledgments}
This work was supported by the Zhejiang Provincial Natural Science Foundation of China (Grant NO. LY23F020025), the Science and Technology Program of Huzhou (Grant NO. 2024GZ09), the Zhejiang Province Leading Geese Plan (Grant NO. 2025C02025), and the National Natural ScienceFoundation of China (Grant NO. 62576292).

\bibliography{aaai2026}

\newpage 
\appendix
\twocolumn[{
  \begin{@twocolumnfalse}
    \section{SD-PSFNet: Sequential and Dynamic Point Spread Function Network for Image Deraining\\Supplementary Material}
  \end{@twocolumnfalse}
}]

\begin{abstract}
In the supplementary materials, we will provide additional experiments and visual analyses to better demonstrate the effectiveness and internal mechanisms of SD-PSFNet. Specifically, to better reveal the internal mechanisms of SD-PSFNet during the training process, we will offer additional analytical perspectives. These analyses include: cross-stage effectiveness analysis under the Sequential Design and Enhanced Feature Fusion mechanism, cross-stage feature information flow analysis, Visualization of PSF Prediction, and Visualization of PSF-Aware Attention. Through these comprehensive analyses, we aim to provide deeper insights into how our proposed architecture functions and why it achieves superior performance.
\end{abstract}

\section{Experiments Supplement}
\begin{figure}[h]
  \centering
  \includegraphics[width=\linewidth]{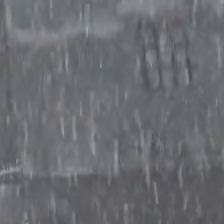}
  \caption{Example images used in additional visualization experiments.}
  \label{fig:1}
\end{figure}

In the supplementary experiments involving visualization, we will select images from the RealRain-1k-L dataset as examples, specifically image \ref{fig:1}, to conduct the related analysis operations.

\begin{figure*}[h]
  \centering
  \includegraphics[width=\linewidth]{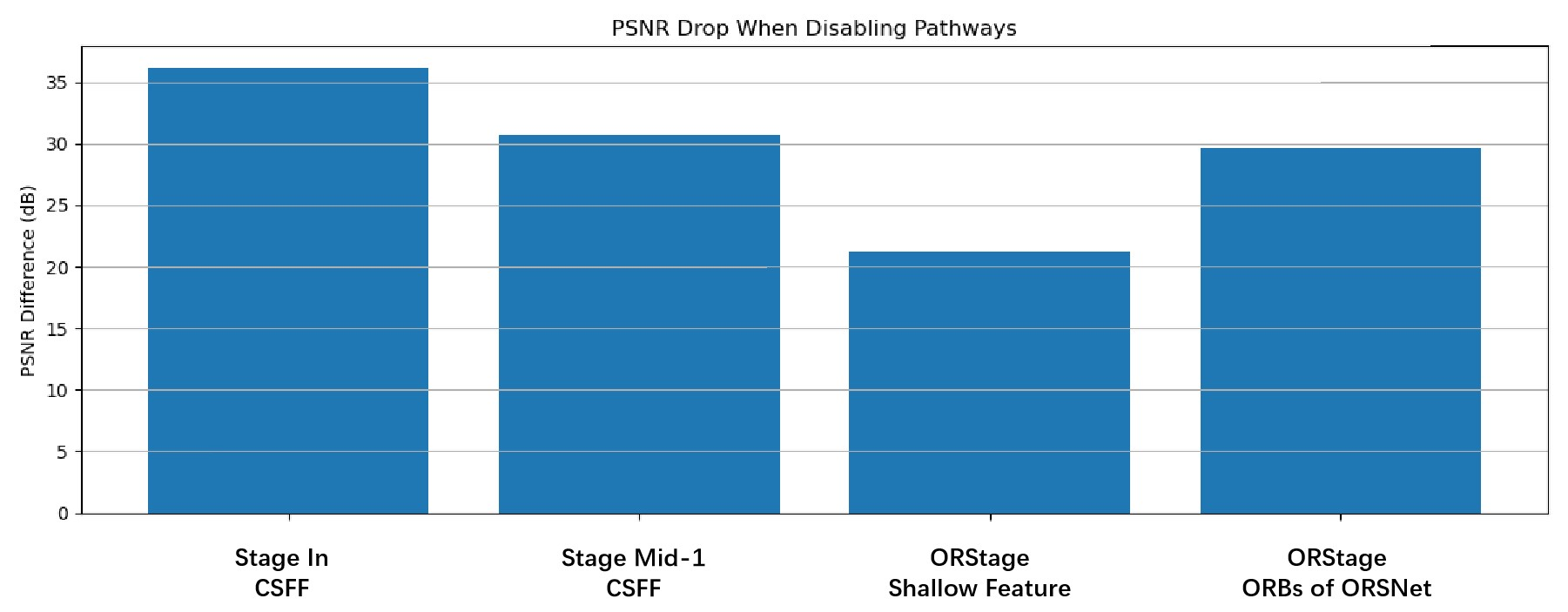}
  \caption{PSNR drop when disabling pathways of gate-based feature fusion between different stages and components.}
  \label{fig:6}
\end{figure*}

\begin{figure*}[h!]
  \centering
  \includegraphics[width=\linewidth]{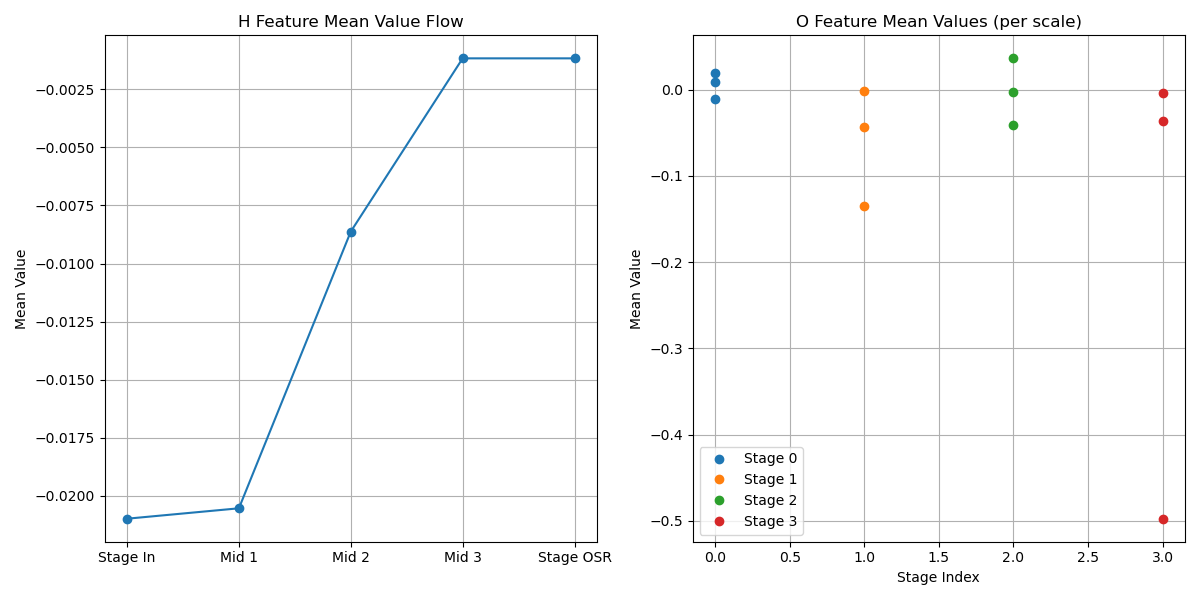}
  \caption{Information Flow Analysis. Stages 0-4 represent Stage In and three Stage Mid phases, demonstrating the numerical transfer of cross-scale features O via CSFF to subsequent stages.}
  \label{fig:4}
\end{figure*}

\subsection{Effectiveness Analysis of Enhanced Feature Fusion}
Enhanced feature fusion is primarily manifested in the cross-stage design of SD-PSFNet. In this section, we explore the effectiveness of gate-based feature fusion between different stages and component paths. As shown in the results in Figure \ref{fig:6}, using the performance trained on RealRain-1k-L in the main text as a baseline, it demonstrates the impact on performance through four methods of disabling these feature transfers under components at different stages. Disabling the feature fusion of the CSFF component at the Stage In stage results in a significant signal drop to approximately 30 dB, indicating its important role in recovery quality. Similarly, the first phase of Stage Mid shows a comparable performance decline. Most critically, disabling feature fusion at the Shallow Feature in ORStage leads to the largest PSNR reduction, as the features from previous training stages cannot be supplemented in the final stage of training. This highlights the importance of feature fusion in the early stages and emphasizes the effectiveness of the gating mechanism used for enhanced feature fusion in SD-PSFNet.

\subsection{Information Flow Analysis}
Figure~\ref{fig:4} illustrates the information flow of two features across different stages through our progressive image restoration network. The left graph tracks the mean value flow of intermediate feature representation \(H\) across processing stages, revealing an important transformation pattern. Here, SD-PSFNet maintains the design described in the main text with three Stage Mid components. Starting from a lower value at the initial stage, we observe minimal change during Stage Mid-1, followed by substantial and continuous growth through Stage Mid-2 and Stage Mid-3, before stabilizing at the ORStage. This trajectory indicates the gradual refinement of advanced feature representations as the model progresses through the restoration pipeline.

The right plot displays the mean values of cross-stage features \( O\) across different scales for each processing stage. The distribution pattern evolves distinctly across the four stages, with Stage 0 (Stage In) values clustered near zero, Stage 1 (Stage Mid 1) showing increased variability, and Stages 2 and 3 (Stage Mid 2 and 3) exhibiting more defined patterns with some values notably decreasing. This scale-dependent feature evolution highlights how our model progressively adapts its internal representations at different spatial frequencies. The systematic changes in both feature types provide empirical evidence of our model's effective hierarchical processing strategy, where earlier stages establish foundational representations that are subsequently refined in later stages to achieve optimal restoration quality.

\begin{figure*}[h!]
  \centering
  \includegraphics[width=\linewidth]{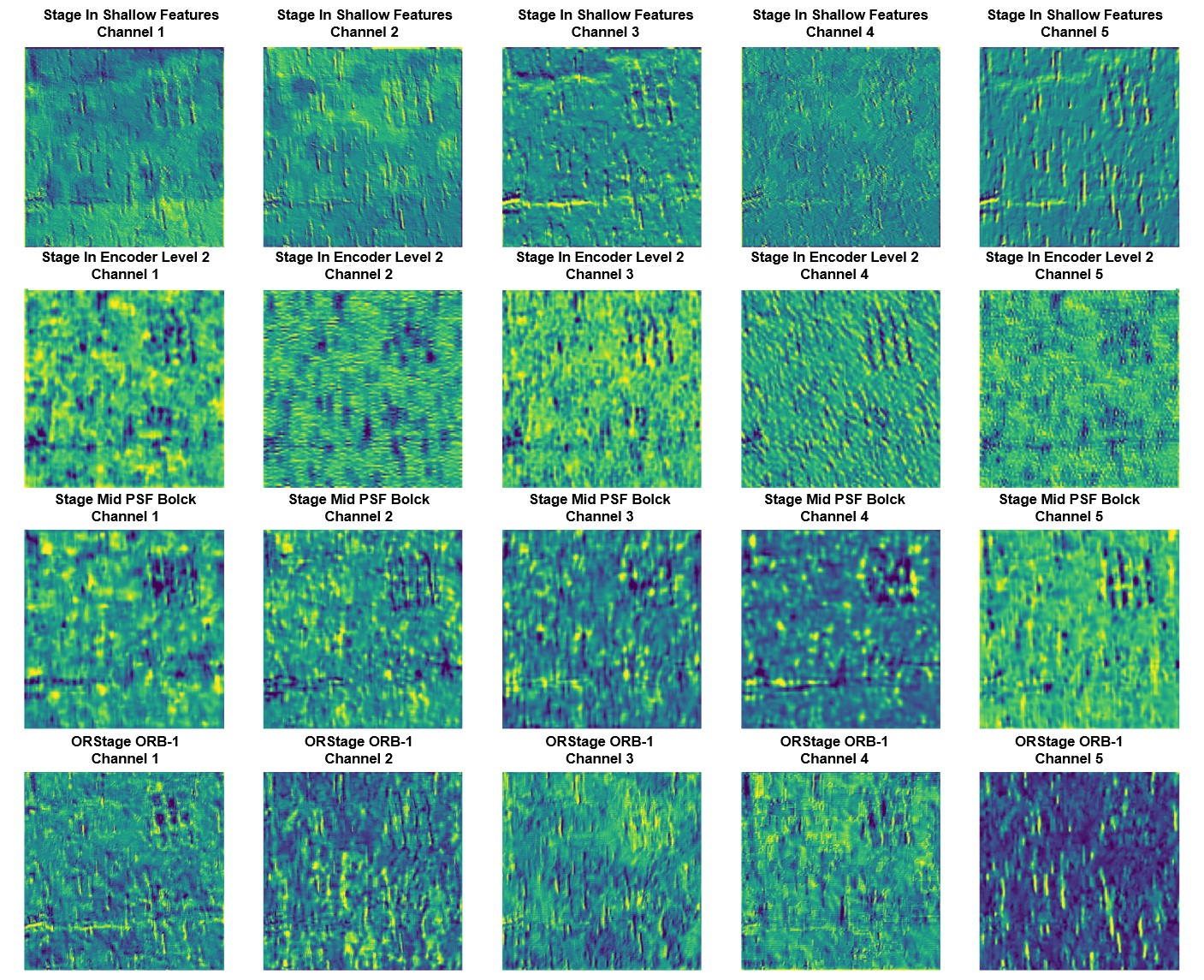}
  \caption{Feature Visualization and Analysis.}
  \label{fig:2}
\end{figure*}

\subsection{Feature Visualization and Analysis}
Figure~\ref{fig:2} displays feature visualizations across multiple stages of our hierarchical image restoration network. The visualization is organized in four rows, each representing different network stages with increasing processing depth.

The top row shows shallow feature channels from the initial stage, capturing predominantly vertical structural patterns and edge information. The second row visualizes features from the second encoder, where more complex textural patterns emerge, with channel 1 particularly highlighting directional wave-like structures. The third row presents Stage Mid decoder PSF block (PSFB) features, where we observe more localized activation patterns that begin to reconstruct finer image details. The bottom row displays specialized OSRNet's ORB-1 (first Original Resolution Block in OStage) features, where each channel demonstrates distinct specialization for different aspects of image restoration.

This hierarchical feature visualization demonstrates how our network progressively processes information: shallow layers capture basic structural elements, intermediate encoder layers extract textural patterns, mid-level decoder layers begin reconstruction, and deeper OSR layers refine specific degradation corrections. The complementary nature of different channels at each stage illustrates our network's ability to simultaneously address various aspects of image degradation through specialized feature representations.

\end{document}